# Robust Subspace Discovery by Block-diagonal Adaptive Locality-constrained Representation

Zhao Zhang[1,2,3], Jiahuan Ren[1], Sheng Li[4], Richang Hong[2,3], Zhengjun Zha[5], Meng Wang[2,3]
[1] School of Computer Science and Technology, Soochow University, Suzhou 215006, China
[2] Key Laboratory of Knowledge Engineering with Big Data (Ministry of Education), Hefei University of Technology
[3] School of Computer Science and Information Engineering, Hefei University of Technology, Hefei, China
[4] Department of Computer Science, University of Georgia, 549 Boyd GSRC, Athens, GA 30602
[5] School of Information Science and Technology, University of Science and Technology of China, Hefei, China
cszzhang@gmail.com, hongrc.hfut@gmail.com, eric.mengwang@gmail.com, sheng.li@uga.edu, zhazj@ustc.edu.cn

## ABSTRACT

We propose a novel and unsupervised representation learning model, i.e., Robust Block-Diagonal Adaptive Locality-constrained Latent Representation (rBDLR). rBDLR is able to recover multi-subspace structures and extract the adaptive locality-preserving salient features jointly. Leveraging on the Frobenius-norm based latent low-rank representation model, rBDLR jointly learns the coding coefficients and salient features, and improves the results by enhancing the robustness to outliers and errors in given data, preserving local information of salient features adaptively and ensuring the block-diagonal structures of the coefficients. To improve the robustness, we perform the latent representation and adaptive weighting in a recovered clean data space. To force the coefficients to be block-diagonal, we perform auto-weighting by minimizing the reconstruction error based on salient features, constrained using a block-diagonal regularizer. This ensures that a strict block-diagonal weight matrix can be obtained and salient features will possess the adaptive locality preserving ability. By minimizing the difference between the coefficient and weights matrices, we can obtain a block-diagonal coefficients matrix and it can also propagate and exchange useful information between salient features and coefficients. Extensive results demonstrate the superiority of rBDLR over other state-of-the-art methods.

## 1 INTRODUCTION

Robust data representation by low-rank coding has been playing an important role for both static or dynamic image sequence (e.g., video) recovery, de-noising and salient representation in areas of multimedia and image processing [7-15]. Real-world data are usually noisy or corrupted, contain mixed structures and high-dimensional features. Low-rank representation algorithm mainly studies how to discover the underlying subspace structures and obtain the compressed representations [1-5][32-50].

To recover the low-rank subspace structures, *Robust Principal Component Analysis* (RPCA) [7] and *Low-Rank Representation* (LRR) [10] are two most representative models. Specifically, LRR and RPCA recover the underlying structures by decomposing data into a recovered low-rank part and a sparse error fitting the noise or corruptions. RPCA only assumes that the data points are drawn from a single low-rank subspace, while LRR studies a more general case that the samples are described using a union of multiple subspaces and the subspace segmentation problem is also considered. Because real-world data usually have the mixed subspace structures in practice, the recovery result of LRR will be potentially more accurate than that of RPCA.

Although RPCA and LRR can recover the subspace structures of data to some extent, they are essentially transductive methods. Thus, they both cannot handle new data efficiently and have to recalculate over all data again when a new sample comes, which will result in high cost. To enable the low-rank methods for joint salient feature extraction, some inductive algorithms have been proposed, such as *Inductive Robust Principal Component Analysis* (IRPCA) [8], *Latent Low-Rank Representation* (LatLRR) [16] and *Joint Low-Rank and Sparse Principal Feature Coding* (LSPFC) [18]. IRPCA addresses the out-of-sample issue by seeking a low-rank projection to extract the low-rank salient features directly, while LSPFC exploits the low-rank and sparse properties of samples jointly and obtains the joint low-rank and sparse salient features [18]. In contrast, LatLRR, as a combination of LRR and IRPCA, improves the performance by integrating latent observations to recover the hidden effects for low-rank basis estimation. More specifically, LatLRR defines the dictionary by employing both observed and unobserved hidden data, which can also resolve the suffered insufficient sampling issue suffered in LRR for more accurate low-rank representation.

Based on discovering the hidden effects, LatLRR has delivered the enhanced recovery and segmentation results, but it still suffers from several problems to degrade the performance and efficiency. First, LatLRR uses the Nuclear-norm to approximate the rank function for low-rank coding, but the Nuclear-norm based optimization involves the *Singular Value Decomposition* (SVD) of matrices, which is usually time-consuming especially for large-scale cases [19]. To improve the efficiency, Frobenius-norm based Latent Low-Rank Representation (FLLRR) [19] was recently proposed, which approximates the rank function using the Frobenius-norm rather than Nuclear-norm. But regularizing the Frobenius-norm on projection and coefficients directly will be sensitive to noise and outliers in data, which may produce the inaccurate representations. Second, both LatLRR and FLLRR fail to encode and preserve the locality structures of samples, so the discovered subspaces may be inaccurate in reality. To preserve the locality structures, *Laplacian Regularized LatLRR* (rLRR) [17] was recently derived, which includes a data-dependent Laplacian regularization over coefficients and salient features into LatLRR. Note that rLRR pre-calculates a graph adjacency matrix from original data to encode the locality, but it cannot ensure the coefficients to be strictly block-diagonal, which is also suffered in LatLRR, and rLRR also suffers from the inefficiency issue due to the SVD process for solving the Nuclear-norm based problem. Moreover, the pre-obtained weights of rLRR cannot be ensured to be joint optimal for subsequent representation and also rLRR cannot ensure the graph Laplacian to be adaptive to different datasets by fixing the neighborhood size. Third, LatLRR, FLLRR and rLRR perform the representation learning in original space that usually contains various noise or even errors, therefore their discovered subspaces may not be accurate as well. The locality of rLRR is also encoded in original input space, which may cause inaccurate similarities due to the negative effects of noise.

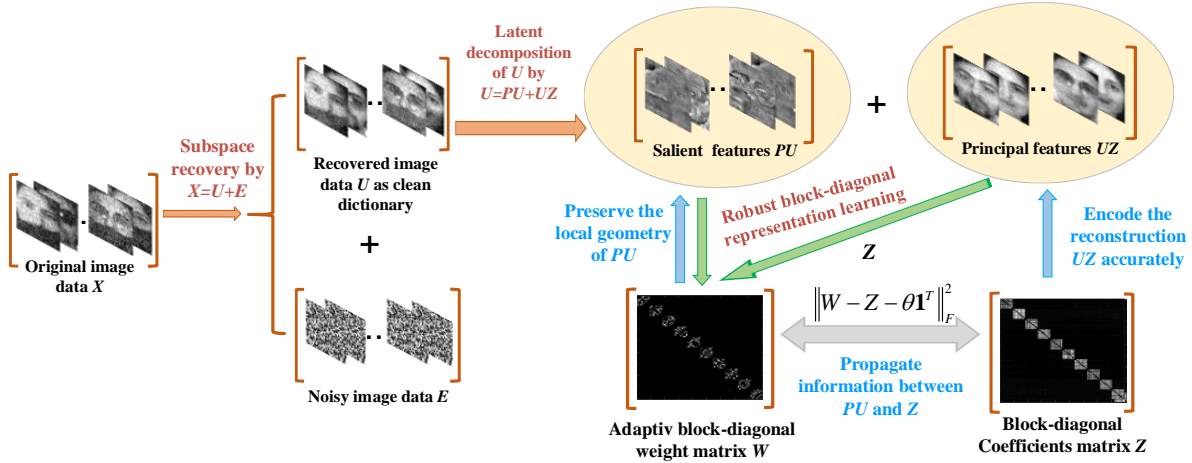

**Fig. 1:** Flow-diagram of our rBDLR, where we show the relations of recovered data, salient features and adaptive coefficients.

In this paper, we mainly propose effective strategies to solve the above drawbacks for more efficient and enhanced subspace recovery. The major contributions are given as follows:

(1) Technically, a novel and effective *Robust Block-Diagonal Adaptive Locality-constrained Latent Representation* model called rBDLR is proposed, which uses the Frobenius-norm for low-rank coding as FLLRR for the efficient representation by avoiding the SVD similarly. To recover the subspaces more accurately, rBDLR encodes the adaptive locality and compute the robust and strict block-diagonal representations. Specifically, the setting of rBDLR integrates the robust strict block-diagonal representation and adaptive locality-based salient feature extraction by robust auto-weighting into a unified representation learning framework.

(2) To enable the coefficients to deliver strict block-diagonal structures for enhancing the representations, rBDLR performs the auto-weighting by minimizing the reconstruction error based on robust adaptive locality-constrained features, constrained by a block-diagonal regularizer over adaptive weights, motivated by *Block-Diagonal Representation* (BDR) [20]. Different from BDR, we did not use an arbitrary matrix for the representation matrix, but use the adaptive locality based regularizer to make the result interpretable, which can also encode the codes more accurately to characterize the subspace structures and pairwise similarities. The adaptive weighting selects the most compact samples to represent data, thus it can make the extracted salient features discriminative and adaptive to different subject classes although the learning manner is unsupervised. Then, rBDLR minimizes the approximation error between the adaptive weight matrix and codes so that the coefficient matrix can have a block-diagonal structure as well. Besides, such an operation can also propagate certain useful information between features and codes.

(3) To improve the robustness to noise and outliers in data, rBDLR performs the block-diagonal representation, salient local feature extraction and adaptive weighting in a recovered clean data space, inspired by the recent *Robust LatLRR* (rLatLRR) [26]. But rLatLRR only performs robust representation learning, which is different from rBDLR. In addition, rLatLRR suffers from the inefficiency issue as LatLRR, cannot preserve the locality of samples and ensure the block-diagonal structures of codes.

The paper is outlined as follows. Section 2 reviews the related work briefly. Section 3 presents rBDLR. Section 4 illustrates the connections between rBDLR and other methods. Section 5 shows the experimental results. Finally, Section 6 concludes the paper.

## 2 RELATED WORK

We briefly review the closely-related LatLRR, FLLRR and BDR.

### 2.1 LatLRR and FLLRR

Given a set of training samples $X = [x_1, x_2, \cdots, x_N] \in \mathbb{R}^{n \times N}$, where $x_i \in \mathbb{R}^n$ is a sample represented by using an *n*-dimensional vector and $N$ denotes the number of samples, LatLRR improves LRR using unobserved hidden data $X_H$ to extend the dictionary and overcome the insufficient sampling problem of LRR. Specifically, LatLRR considers the following low-rank coding problem:

$$\min_{Z} rank(Z), s.t. X_O = [X_O, X_H]Z, \quad (1)$$

where $rank(\cdot)$ is the rank function and $X_O$ is the observed data. If $X_O$ and $X_H$ are all sampled from the same collection of low-rank subspaces, by using the Nuclear-norm to approximate the rank function and using L$_1$-norm on the sparse error $E$, LatLRR can recover the hidden effects by minimizing

$$\min_{Z,P,E} \|Z\|_* + \|P\|_* + \lambda \|E\|_1, s.t. X = XZ + PX + E, \quad (2)$$

where $\|Z\|_*$ is the Nuclear-norm of $Z$ [16-17][21], i.e., the sum of its singular values, $XZ$ and $PX$ are principal features and salient features respectively, and $\lambda$ is a positive weighting factor.

Note that LatLRR uses the Nuclear-norm constraint on $Z$ and $P$, so the time-consuming SVD process will be involved. Note that Frobenius-norm $\|\cdot\|_F$ can also be used as the convex surrogate of rank function [19]. Besides, the optimization of Frobenius-norm is more efficient compared with Nuclear-norm. The Frobenius-norm based objective function of FLLRR is defined as

$$\min_{Z,P,E} \|Z\|_F + \|P\|_F + \lambda \|E\|_1, s.t. X = XZ + PX + E. \quad (3)$$

## 2.2 Block Diagonal Representation (BDR)

Existing LRR based methods [9][16-18] usually approximate the block-diagonal representation matrix by using different structure priors, such as sparsity and low-rankness, which are all indirect methods. BDR uses a block-diagonal matrix induced regularizer for pursuing a block-diagonal representation directly. Based on the block-diagonal regularizer, recent BDR solves the subspace clustering problem by using the block-diagonal structure prior in ideal case. The $k$-block diagonal regularizer is defined as

$$\|B\|_{\underline{k}} = \sum_{i=m-k+1}^{m} \lambda_i(L_B), s.t. B \geq 0, B^T = B, \quad (4)$$

where $\lambda_i(L_B), i=1,\ldots,m$ are eigenvalues of $L_B$ in the decreasing order, $B$ is an arbitrarily defined representation matrix and $L_B$ is the Laplacian matrix over the representation matrix $B$:

$$L_B = Diag(B\mathbf{1}) - B, \quad (5)$$

where $Diag(B\mathbf{1})$ is a diagonal matrix with its $i$-th diagonal entry being $(B\mathbf{1})_i$, $\mathbf{1} \in \mathbb{R}^{N \times 1}$ denotes a column vector of all ones and $B\mathbf{1}$ is the product between the matrix $B$ and vector $\mathbf{1}$.

Based on the $k$-block diagonal regularizer, BDR considers the following problem for handling data with noise:

$$\min_{B} \frac{1}{2}\|X - XB\|_F^2 + \gamma\|B\|_{\underline{k}}, s.t. diag(B) = 0, B \geq 0, B = B^T, \quad (6)$$

where $\|\cdot\|_{\underline{k}}$ is a $k$-block diagonal matrix based regularizer and the representation matrix $B$ should be nonnegative and symmetric, which are necessary to define the block-diagonal regularizer. But the restrictions on $B$ may limit its representation power, so BDR alleviates it by introducing an intermediate term $\|Z - B\|_F^2$:

$$\min_{Z,B} \frac{1}{2}\|X - XZ\|_F^2 + \frac{\lambda}{2}\|Z - B\|_F^2 + \gamma\|B\|_{\underline{k}}. \quad (7)$$
$$s.t. \ diag(B) = 0, B \geq 0, B = B^T$$

The above problems in Eqs.(6) and (7) are equivalent as $\lambda > 0$ is sufficiently large. $\|Z - B\|_F^2$ can make the problem for updating $Z$ and $B$ strongly convex and hence the solutions are unique and stable. But the matrix $B$ is arbitrarily defined, which has no specific meanings. Although minimizing $\|Z - B\|_F^2$ forces $Z$ to deliver the block-diagonal structures by approximating it using $B$, the obtained $Z$ may not be accurate to characterize the subspace structures and similarities, which may also cause the overfitting in reality, since the constraint $\|Z - B\|_F^2$ is too strong and hard.

## 3 PROPOSED FORMULATION

### 3.1 Objective Function

We introduce the formulation of rBDLR algorithm in this section. rBDLR mainly improves the representation ability from twofold. First, it defines a block-diagonal regularizer with clear definition and interpretability on the representation matrix so that the recovered subspaces are more accurate. Specifically, rBDLR first calculates an adaptive weight matrix $W$ using the salient features $PX$ by minimizing the reconstruction error over $PX$, and then uses it to define the representation matrix, which is different from BDR that applies an arbitrary matrix for the representation matrix. To avoid the overfitting issue and relax the constraint $\|W - Z\|_F^2$, we use a flexible approximation error $\|W - Z - \theta\mathbf{1}^T\|_F^2$, where $\theta \in \mathbb{R}^{N \times 1}$ denotes a bias. By minimizing $\|W - Z - \theta\mathbf{1}^T\|_F^2$, i.e., approximating $W$ using $Z + \theta\mathbf{1}^T$, one can achieve a coefficient matrix potentially with clear block-diagonal structures, as $\theta\mathbf{1}^T$ will contain some mixed signs and noise in $W$. Thus, the initial problem of rBDLR in noise-free case can be defined as

$$\min_{Z,P,W,\theta} \|Z\|_F^2 + \|P\|_F^2 + \alpha\|W - Z - \theta\mathbf{1}^T\|_F^2 + \beta(\|PX - PXW\|_F^2 + \|W\|_{\underline{k}}).$$
$$s.t. \ X = XZ + PX, diag(W) = 0, W = W^T, W \geq 0$$
(8)

By imposing the block-diagonal constraint $\|\cdot\|_{\underline{k}}$ on $W$, one can obtain a strict block-diagonal auto-weighting matrix $W$. By the minimization of $\|PX - PXW\|_F^2$, extracted salient features $PX$ will have the adaptive locality preserving power, and moreover will be joint discriminative and adaptive to different subject classes although the learning process is unsupervised. By minimizing $\|W - Z - \theta\mathbf{1}^T\|_F^2$, one can explicitly ensure the coefficient matrix $Z$ to have block-diagonal structures. Note that $W$ is obtained from the salient features $PX$, and both $Z$ and $PX$ are the low-rank coefficients and salient features of $X$ respectively. Thus, $Z$ and $PX$ should share certain important local geometrical information. So, minimizing $\|W - Z - \theta\mathbf{1}^T\|_F^2$ to make $Z$ infinitely close to $W$ will be reasonable. Besides, minimizing $\|W - Z - \theta\mathbf{1}^T\|_F^2$ can also propagate some hidden information from salient features $PX$ to coefficients $Z$, which can make the decomposition more accurate.

Since real application data often have various noise and errors, the obtained representation and weights may be inaccurate in reality. To tackle this issue, rBDLR considers the joint robust representation and weighting. Specifically, rBDLR first de-noises $X$ by correcting errors similarly as [26] using L$_{2,1}$-norm [21] and then performs representation and auto-weighting in a recovered clean space $X$-$E$, which leads to the following objective function:

$$\min_{Z,P,E,W,\theta} \|Z\|_F^2 + \|P\|_F^2 + \|A_+ - A_-W\|_F^2 + \beta\|W\|_{\underline{k}} + \gamma\|E\|_{2,1}.$$
$$s.t. \ (X - E) = (X - E)Z + P(X - E), diag(W) = 0, W = W^T, W \geq 0$$
(9)

where $diag(W) = 0$ can avoid the trivial solution $W=I$, and $A_+$ and $A_-$ are two auxiliary matrices that are defined as

$$A_+ = \begin{pmatrix} \sqrt{\alpha}(Z+\theta\mathbf{1}^T) \\ \sqrt{\beta}P(X-E) \end{pmatrix}, A_- = \begin{pmatrix} \sqrt{\alpha}I \\ \sqrt{\beta}P(X-E) \end{pmatrix}. \quad (10)$$

Note that the flow-diagram of our rBDLR is illustrated in Fig.1, where the relations of the recovered data, salient features and adaptive coefficients are shown clearly. The problem of rBDLR can be performed alternately between the following two steps:

**(1) Robust block-diagonal subspace discovery**
We first fix the block-diagonal representation matrix $W$ for the robust subspace recovery and feature extraction. When $W$ is known, we have the following problem from Eq.(9):

$$\min_{Z,P,E,\theta} \|Z\|_F^2 + \|P\|_F^2 + \alpha\|Z+\theta\mathbf{1}^T - W\|_F^2$$
$$+ \beta\|P(X-E) - P(X-E)W\|_F^2 + \gamma\|E\|_{2,1}. \quad (11)$$
$$s.t. \ (X-E) = (X-E)Z + P(X-E)$$

**(2) Adaptive locality based block-diagonal representation:**
When the projection $P$, representation coefficients $Z$ and sparse error $E$ are given, we can seek the block-diagonal representation matrix $W$ from the following simplified problem:

$$\min_W \|A_+ - A_- W\|_F^2 + \beta\|W\|_{\underline{k}}, \ s.t. \ diag(W)=0, W=W^T, W \geq 0. \quad (12)$$

Next, we describe the optimization procedures of our rBDLR.

### 3.2 Optimization

It is easy to check from Eq.(9) that the involved variables depend on each other. Following the common procedures, we solve Eq.(9) by updating the variables alternately and use inexact Augmented Lagrange Multiplier method (ALM) [22] for efficiency. Let $U$ be an auxiliary matrix, we first convert the problem into

$$\min_{Z,P,E,W,\theta} \|Z\|_F^2 + \|P\|_F^2 + \|F_+ - F_- W\|_F^2 + \beta\|W\|_{\underline{k}} + \gamma\|E\|_{2,1}$$
$$s.t. \ U = UZ + PU, U = X-E, diag(W)=0, W=W^T, W \geq 0 \quad , (13)$$

where $F_+ = \left(\sqrt{\alpha}(Z+\theta\mathbf{1}^T)^T, \sqrt{\beta}U^T P^T\right)^T$ and $F_+ = \left(\sqrt{\alpha}I, \sqrt{\beta}U^T P^T\right)^T$, and the corresponding Lagrangian function is defined as

$$\wp = \|Z\|_F^2 + \|P\|_F^2 + \|F_+ - F_- W\|_F^2 + \beta\|W\|_{\underline{k}}$$
$$+ \gamma\|E\|_{2,1} + tr(Y_1^T(U - UZ - PU)) + tr(Y_2^T(U-X-E)), \quad (14)$$
$$+ \frac{\mu}{2}\left(\|U - UZ - PU\|_F^2 + \|U - X - E\|_F^2\right)$$

w.r.t. $diag(W)=0, W=W^T, W\geq 0$, where $Y_1$ and $Y_2$ are Lagrange multipliers, and $\mu$ is a positive factor. By inexact ALM, rBDLR updates the variables by solving the Lagrange function $\wp$ :

$$(P_{t+1}, Z_{t+1}, W_{t+1}, E_{t+1}, \theta_{t+1}) = \arg\min_{P,Z,W,E} \wp_t(P, Z_t, W_t, E_t, \theta_t)$$
$$Y_1^{t+1} = Y_1^t + \mu_t(U_{t+1} - P_{t+1}U_{t+1} - U_{t+1}Z_{t+1})$$
$$Y_2^{t+1} = Y_2^t + \mu_t(U_{t+1} - X - E_{t+1}), \mu_{t+1} = \min(\eta\mu_t, \max_\mu) \quad . \quad (15)$$

Next, we will show the procedures of updating the variables.

**1) Fix $W$ and $E$, update $Z$, $P$ and $\theta$ :**
When the representation matrix $W$ and error $E$ are known, we can update $P$ and $Z$ from the following sub-problem:

$$\min_{Z,P,\theta} \|Z\|_F^2 + \|P\|_F^2 + \left\| \begin{pmatrix} \sqrt{\alpha}(Z+\theta\mathbf{1}^T) \\ \sqrt{\beta}PU \end{pmatrix} - \begin{pmatrix} \sqrt{\alpha}I \\ \sqrt{\beta}PU \end{pmatrix} W \right\|_F^2. \quad (16)$$
$$+ tr(Y_1^T(U-UZ-PU)) + \frac{\mu}{2}\|U-UZ-PU\|_F^2$$

By taking the derivatives w.r.t. $Z$ and $P$, and zeroing the derivatives respectively, we can infer $Z_{t+1}$ and $P_{t+1}$ as follows:

$$Z_{t+1} = (2I + 2\alpha I + \mu_t U_t^T U_t)^{-1} \Psi_t, \quad (17)$$

$$P_{t+1} = (Y_1^t U_t^T + \mu_t U_t U_t^T - \mu_t U_t Z_t U_t^T)(\Gamma_t + 2I)^{-1}, \quad (18)$$

where $\Gamma_t = (2\beta + \mu_t)U_t U_t^T - 4\beta U_t W_t U_t^T + 2\beta U_t W_t W_t^T U_t^T$ and $\Psi_t = 2\alpha W_t + U_t^T Y_1^t + \mu_t U_t^T U_t - \mu_t U_t^T P_{t+1} U_t - 2\alpha \theta_t \mathbf{1}^T$.

Let $J$ denote the objective function of Eq.(16), by taking the derivative of $J$ w.r.t. $\theta$, we can easily update $\theta$ as

$$\partial J/\partial\theta = \theta\mathbf{1}^T\mathbf{1} + Z\mathbf{1} - W\mathbf{1} = 0 \Rightarrow \theta_{t+1} = (W_t\mathbf{1} - Z_{t+1}\mathbf{1})/N. \quad (19)$$

**2) Fix $Z, P, W, \theta$, update $E$ and $U$:**
By removing the terms irrelevant to $E$ and $U$ from Eq.(14), we have the following reduced formulation:

$$E_{t+1} = \arg\min_E \frac{\gamma}{\mu_t}\|E\|_{2,1} + \frac{1}{2}\left\|E - (U_t - X + (1/\mu_k)Y_2^t)\right\|_F^2. \quad (20)$$

According to [10][22], let $\Theta = U_t - X + (1/\mu_k)Y_2^t$ and denote by $\Theta_{:,i}$ the $i$-th column of $\Theta$, the $i$-th column $E_{t+1}^{:,i}$ of solution $E_{t+1}$ at the $(t+1)$-th iteration can be obtained as

$$E_{k+1}^{:,i} = \begin{cases} \dfrac{\|\Theta_{:,i}\|_2 - (\gamma/\mu_k)}{\|\Theta_{:,i}\|_2}\Phi_{:,i}^E & \text{if } (\gamma/\mu_k) < \|\Theta_{:,i}\|_2 \\ 0 & \text{otherwise} \end{cases}. \quad (21)$$

After the sparse error $E_{t+1}$ is updated at each iteration, we can update the auxiliary matrix $U$ by $U_{t+1} = X - E_{t+1}$.

**3) Fix others, update the block-diagonal matrix $W$:**
When the projection $P$, coefficients $Z$ and error $E$ are known, we can update the block-diagonal matrix $W$. By removing the terms irrelevant to $W$ from Eq. (14), we have the following problem:

$$\min_W \left\| \begin{pmatrix} \sqrt{\alpha}(Z+\theta\mathbf{1}^T) \\ \sqrt{\beta}PU \end{pmatrix} - \begin{pmatrix} \sqrt{\alpha}I \\ \sqrt{\beta}PU \end{pmatrix}W \right\| + \beta\|W\|_{\underline{k}}, \ s.t. \ \begin{matrix}diag(W)=0 \\ W=W^T, W\geq 0\end{matrix}. \quad (22)$$

According to [20], the Laplacian matrix $L_W$ of affinity matrix $W$ is defined as $L_W = Diag(W\mathbf{1}) - W$, where $Diag(W\mathbf{1})$ denotes a diagonal matrix with its $i$-th entry being $(W\mathbf{1})_i$. If the Laplacian matrix $L_W$ is positive semi-definite, we define $L_W \succeq 0$. Then, we can reformulate $\|W\|_{\underline{k}}$ as a convex problem:

$$\|W\|_{\underline{k}} = \sum_{i=m-k+1}^m \lambda_i(L_W) = \min_M \langle L_W, M\rangle$$
$$s.t. \ 0 \preceq M \preceq I, Tr(M) = k \quad . \quad (23)$$

Then, we have the following equivalent problem for Eq.(22):

$$\min_{W,M} \left\| \begin{pmatrix} \sqrt{\alpha}Z_t \\ \sqrt{\beta}PU \end{pmatrix} - \begin{pmatrix} \sqrt{\alpha}I \\ \sqrt{\beta}PU \end{pmatrix}W \right\|_F^2 + \beta\langle Diag(W\mathbf{1}) - W, M\rangle. \quad (24)$$
$$s.t. \ W \geq 0, W^T = W, 0\preceq M \preceq I, Tr(M) = k$$

| **Algorithm 1**: Optimization procedures of our rBDLR |
|---|
| **Inputs:** Training dataset $X$, tuning parameters $\alpha, \beta, \gamma$.<br>**Initialization:** $P_0=0$, $E_0=0$, $W_0=0$, $U_0=X$, $\mu_0=10^{-6}$, $\max_\mu = 10^{10}$, $\theta_0 = 0, Y_1^0 = 0, Y_2^0 = 0, \varepsilon = 10^{-7}, \mu_0 = 10^{-6}, \max_\mu = 10^{10}, \eta = 1.12, t = 0$.<br>**While** *not converged* **do**<br>1. Fix others, update the coefficients $Z_{t+1}$ by Eq.(17);<br>2. Fix others, update the projection $P_{t+1}$ by Eq.(18);<br>3. Fix others, update bias term $\theta_{t+1}$ by Eq.(19);<br>4. Fix others, update the sparse error $E_{t+1}$ by Eq.(20) and then update $U$ by $U_{t+1}=X - E_{t+1}$;<br>5. Fix others, update the block-diagonal $W_{t+1}$ by Eq.(26);<br>6. Update the multipliers $Y_1$ and $Y_2$ by Eq.(15);<br>7. Update the parameter $\mu$ with $\mu_{t+1} = \min(\eta\mu_t, \max_\mu)$;<br>8. Convergence check: if max ($\|U_{t+1} - U_{t+1}Z_{t+1} - P_{t+1}U_{t+1}\|_\infty$, $\|U_{t+1} - X - E_{t+1}\|_\infty < \varepsilon$, stop; else $t = t+1$.<br>**End while**<br>**Outputs:** $Z^* \leftarrow Z_{t+1}, P^* \leftarrow P_{t+1}, W^* \leftarrow W_{t+1}$. |

By taking the derivative *w.r.t.* M, we can update $M_{t+1}$ as

$$M_{t+1} = \arg\min_M \langle Diag(W_t\mathbf{1}) - W_t, M \rangle \\ s.t. \ 0 \preceq M \preceq I, Tr(M) = k. \quad (25)$$

Finally, we update the weights $W_{t+1}$ by the following rule:

$$W_{t+1} = \left(2A_-^{t+1T}A_-^{t+1}\right)^{-1}\left(2\Xi_{t+1} - \beta\left(diag(M_{t+1})\mathbf{1}^T - M_{t+1}\right)\right), \quad (26)$$

where $\Xi_{t+1} = A_-^{t+1T}A_+^{t+1}$ and the auxiliary matrices $A_+$ and $A_-$ at the (*t+1*)-th iteration are defined as follows:

$$A_+^{t+1} = \begin{pmatrix} \sqrt{\alpha}(Z_{t+1} + \theta_{t+1}\mathbf{1}^T) \\ \sqrt{\beta}P_{t+1}(X - E_{t+1}) \end{pmatrix}, A_-^{t+1} = \begin{pmatrix} \sqrt{\alpha}I \\ \sqrt{\beta}P_{t+1}(X - E_{t+1}) \end{pmatrix}. \quad (27)$$

For the complete presentation of optimization, we summarize the procedures in Algorithm 1. About the convergence of rBDLR, firstly the convergence property of inexact ALM has been well studied as the number of blocks is at most two [7]. But it is still not easy to prove that the solution of models converge to global optimum theoretically when having over two blocks. We also approach our model by solving the blocks alternately, which are easily solvable at each iteration, similarly as IRPCA, LRR and LatLRR, etc. The time complexity of rBDLR is the same as that of FLLRR. The value of *k* in the *k*-block-diagonal regularizer is set to the number of subject classes in the following simulations.

## 4 RELATIONSHIP ANALYSIS

**Connection between FLLRR and rBDLR.** Recall the objective function of rBDLR in Eq. (9), if $\alpha=0$, $\beta=0$, and the constraints on *W* are removed, we have the following reduced problem:

$$\min_{Z,P,E} \|Z\|_F^2 + \|P\|_F^2 + \gamma\|E\|_{2,1}, s.t. (X - E) = (X - E)Z + P(X - E). \quad (28)$$

By comparing the above formulation with that of FLLRR in Eq.(3), the two problems are very similar in form. But note that although FLLRR and rBDLR can also obtain a coefficient matrix *Z* and a projection *P*, and correct the sparse error *E* in the data at the same time, their mechanisms are still different. Specifically, FLLRR clearly seeks *Z* and *P* from *X*, while rBDLR performs in the recovered space *X-E*. Thus, FLLRR is a special case of rBDLR.

**Connection between BDR and rBDLR.** We first express the problem of BDR in noiseless case as

$$\min_{Z,B} \frac{\lambda}{2}\|Z - B\|_F^2 + \gamma\|B\|_{\boxed{k}}, s.t. \ X = XZ, diag(B) = 0, B \geq 0, B = B^T. \quad (29)$$

For the objective function of rBDLR in Eq. (9), if we remove the Frobenius-norm regularization $\|Z\|_F$ and $\|P\|_F$, we have

$$\min_{Z,P,W} \alpha\|W - Z - \theta\mathbf{1}^T\|_F^2 + \beta\left(\|PX - PXW\|_F^2 + \|W\|_{\boxed{k}}\right). \quad (30) \\ s.t. \ X = XZ + PX, diag(W) = 0, W = W^T, W \geq 0$$

By comparing the problems, we find that BDR minimizes the reconstruction error for learning *Z*, while rBDLR performs latent representation for jointly learning codes and adaptive locality-preserving salient features. Besides, BDR uses a hard constraint $\|Z - B\|_F^2$ and defines a block-diagonal representation matrix as an arbitrary matrix, while rBDLR uses a flexible constraint and unifies the auto-weighting to calculate the interpretable adaptive weights for block-diagonal representation.

## 5 SIMULATION RESULTS AND ANALYSIS

We show the visual and quantitative comparison results of our rBDLR with those of several related algorithms, i.e., RPCA [7], LRR [10], IRPCA [8], I-LSPFC [18], rLRR [17], LatLRR [16], rLatLRR [26], FLLRR [19], BDR [20], *Isoprojection* (IsoP) [29] and *Locality Preserving Projections* (LPP) [28] for feature extraction and classification. Five public databases are evaluated, including two face recognition databases: UMIST [23] and AR [31], two object databases: COIL100 [24] and ETH80 [25], one handwriting database: USPS [27]. For data classification, samples are divided into a training set and a test set, where the training set is to learn a recovered subspace, and the test set is to evaluate the accuracy. Following the common practice, we resize all the face and object images into $32 \times 32$ pixels, which corresponds to a 1024-D data point in vector space. For fair comparison, all parameters of each method are chosen carefully. We perform all simulations on a PC with Intel (R) Core (TM) i7-7700 CPU @ 3.6 GHz 8G.

### 5.1 Investigation of Parameter Selections

Since the parameter selection is still an open problem, a heuristic way is to select the most important ones. ETH80 database is used and we use the features of [15]. The ETH80 database has 8 big categories. In each big category, 10 subcategories are included, each of which has 41 images. That is, it contains 3280 images of 80 objects. In this paper, we consider an eight-class categorization problem, i.e., each of the eight big categories is treated as a single class. We randomly select 4 images from each class as training set and test on the rest. rBDLR has three parameters, so we aim to fix one of them and explore the effects of other two by grid search. To see the effects of different parameters, we first fix $\alpha=1$, then tune $\beta$ and $\gamma$ by grid search from the candidate set $[10^{-7}, 10^{-5}, ..., 10^7]$ as shown in Fig.2(a), from which we find that rBDLR with $10^{-7} \leq \beta \leq 0.1$ can deliver good results. Then we fix $\beta=10^{-5}$ to observe the effects of $\alpha$ and $\gamma$ in Fig.2(b), from which we see that rBDLR with $\alpha \geq 10^{-5}$ obtains stable and good results. Finally, we fix $\gamma=10^{-2}$ to investigate the effects of $\alpha$ and $\beta$ in Fig.2(c), and we see that rBDLR also performs well in most cases. Note that similar findings are obtained from other datasets, thus we can simply select $\alpha \geq 10^{-5}, \beta \leq 10^{-1}$, since rBDLR is robust to $\gamma$.

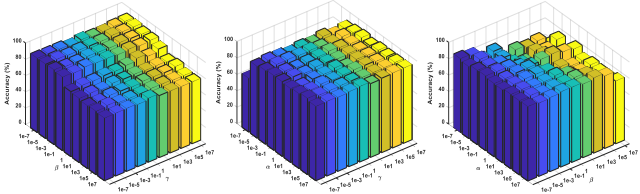

**Figure 2: Parameter sensitivity analysis of rBDLR on ETH80, where (a) fix $\alpha$ to tune $\beta$ and $\gamma$; (b) fix $\beta$ to tune $\alpha$ and $\gamma$; (c) fix $\gamma$ to tune $\alpha$ and $\beta$.**

## 5.2 Convergence Results

We present some convergence results of rBDLR on four image databases, i.e., UMIST, ETH80, COIL100 and USPS, as examples. For each database, we select 10 images per class for training and the averaged convergence results over 10 splits are shown in Fig.3. We find that our rBDLR often converges with the number of iterations ranging from 30 to 80 in most cases.

## 5.3 Visual Image Analysis by Visualization

**Visualization of coefficients Z.** In this study, we follow [16] to construct 10 independent subspaces $\{S_i\}_{i=1}^{10}$ whose bases $\{H_i\}_{i=1}^{10}$ are obtained by $H_{i+1} = GH_i, 1 \leq i \leq 9$, where $G$ is a random rotation and $H_i$ is a random column orthogonal matrix whose dimension is $200 \times 10$ and each subspace has a fixed rank of 10. Then, we construct a data matrix $X = [X_1, X_2, \dots X_{10}] \times \mathbb{R}^{200 \times 10}$ by sampling 9 (that is smaller than the rank of subspace) data vectors from each subspace by $X_i = H_i C_i, 1 \leq i \leq 9$ where $C_i$ is a $10 \times 9$ i.i.d. $N(0,1)$ data matrix. Then, we use this artificial data matrix for coding by LatLRR, FLLRR, BDR and rBDLR. The visualization results of coefficients are shown in Fig. 4. We find that: (1) the coefficients matrix $Z$ of each algorithm have block-diagonal structures. But compared with those of BDR and our rBDLR, the coefficients of LatLRR and FLLRR in the off-diagonal are noisy and contains some wrong inter-class connections. Due to the deployment of a block-diagonal regularizer, a strict block-diagonal coefficients matrix $Z$ is delivered for BDR and rBDLR, but the connectivity within each block of the codes of BDR is worse than rBDLR. This improvement can be attributed to the adaptive weighting with a block-diagonal regularizer to approximate the coefficients.

**Visualization of recovered features XZ.** We then evaluate rBDLR by visualizing the recovered faces $XZ$, that is, principal features. For a face data matrix $X$, rBDLR firstly de-noises it and decomposes it into the principal features $XZ$ and salient features $PX$. In this study, AR database is used, since it has more facial variations, including illumination changes, face expressions, and occlusions by sunglasses and scarf. Random Gaussian noise with variance=500 is added into the face images to examine the robust properties of each method for face recovery. Some original faces, noisy faces and recovered features are shown in Fig.5. We see that the recovery process of rBDLR can remove the occlusions in the face images, and the recovered images of FLLRR, LatLRR, rLRR, BDR contain more noise than those of our rBDLR.

## 5.4 Application to Image Recognition

We evaluate each algorithm for image recognition on the UMIST, COIL100, ETH80 and USPS databases. The results of rBDLR are compared with those of IRPCA, I-LSPFC, LatLRR, rLatLRR, rLRR, FLLRR, IsoP and LPP. Note that all compared methods can obtain a projection for feature extraction. For each new data $x_{new}$, we obtain features by embedding $x_{new}$ onto the projection vectors. Then, one-Nearest-Neighbor (1NN) classifier is used on extracted features for classification. Some examples are shown in Fig.6. For quantitative recognition, the results of each method are averaged based on 10 random splits of training/test samples.

**Recognition results on the UMIST database.** This database contains 1012 images from 20 different individuals that change the poses from profile to frontal views. We randomly select 4, 6, 8 and 10 images from each person as training set and test on the rest. The recognition result of each method are shown in Table 1. We find that the result of each approach can be improved when

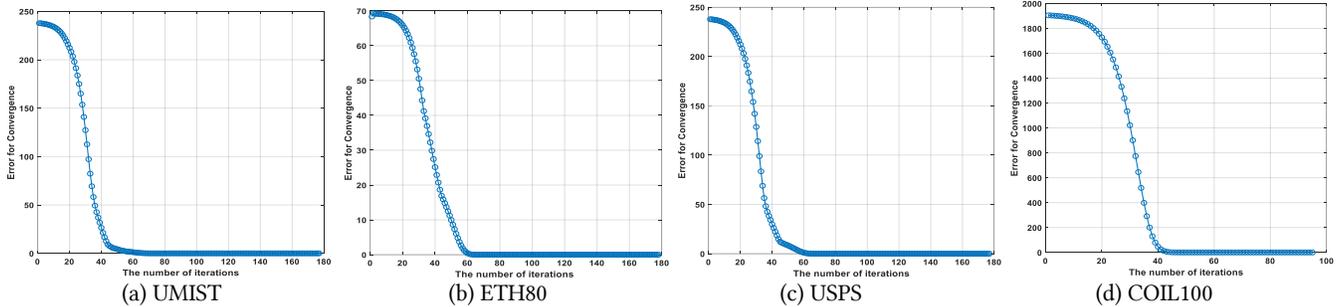

(a) UMIST  (b) ETH80  (c) USPS  (d) COIL100

**Figure 3: Convergence results of our rBDLR algorithm on four real-world image databases.**

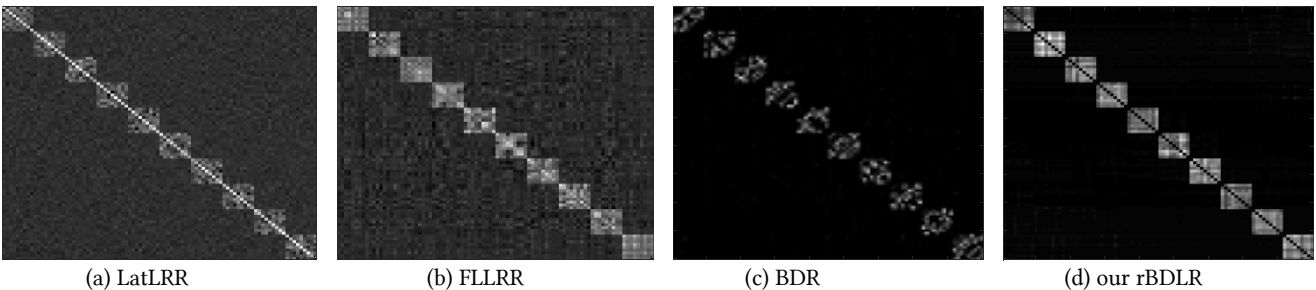

(a) LatLRR  (b) FLLRR  (c) BDR  (d) our rBDLR

**Figure 4: Visual comparison of the representation coefficients matrix Z of LatLRR, FLLRR, BDR and our rBDLR.**

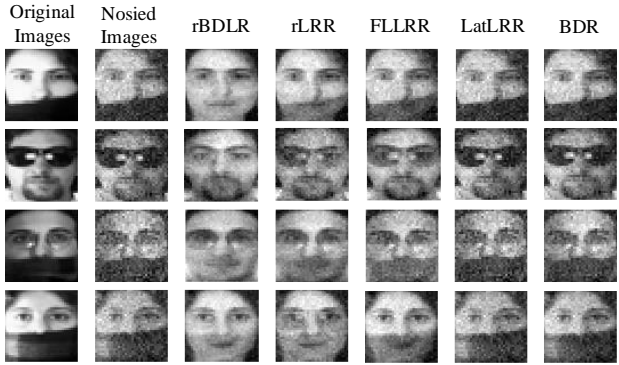

Figure 5: Results of rLRR, FLLRR, LatLRR, BDR and rBDLR.

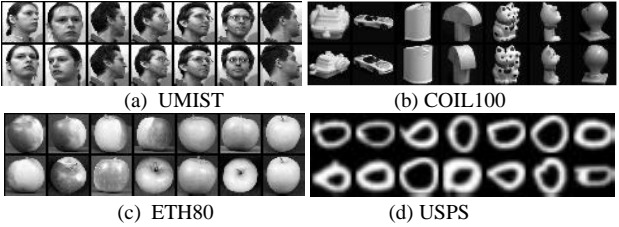

(a) UMIST  (b) COIL100
(c) ETH80  (d) USPS

Figure 6: Image examples of the four evaluated databases.

Table 1: Averaged face recognition rates on UMIST.

| Evaluated Methods | 4 train Acc. (%) | 6 train Acc. (%) | 8 train Acc. (%) | 10 train Acc. (%) |
|---|---|---|---|---|
| LPP | 82.32 | 85.87 | 88.03 | 92.61 |
| IsoP | 82.76 | 86.32 | 91.08 | 94.67 |
| IRPCA | 85.94 | 89.25 | 93.06 | 93.86 |
| I-LSPFC | 87.23 | 89.91 | 94.61 | 95.05 |
| LatLRR | 84.13 | 85.22 | 87.33 | 94.26 |
| rLatLRR | 85.51 | 88.31 | 92.61 | 95.07 |
| FLLRR | 88.48 | 90.14 | 93.66 | 95.61 |
| rLRR | 85.22 | 87.24 | 89.56 | 95.44 |
| **rBDLR** | **89.68** | **92.94** | **96.6** | **97.21** |

Table 2: Averaged object recognition rates on COIL100.

| Evaluated Methods | 5 train Acc. (%) | 10 train Acc. (%) | 15 train Acc. (%) | 20 train Acc. (%) |
|---|---|---|---|---|
| LPP | 57.99 | 67.58 | 77.79 | 83.06 |
| IsoP | 52.11 | 62.10 | 73.56 | 78.32 |
| IRPCA | 54.21 | 64.03 | 75.46 | 81.71 |
| I-LSPFC | 56.21 | 66.32 | 78.82 | 85.21 |
| LatLRR | 54.65 | 65.49 | 76.78 | 79.84 |
| rLatLRR | 70.16 | 79.92 | 84.09 | 87.61 |
| FLLRR | 71.37 | 79.40 | 86.23 | 88.41 |
| rLRR | 56.33 | 67.88 | 79.37 | 82.46 |
| **rBDLR** | **73.12** | **83.53** | **89.42** | **90.23** |

Table 3: Averaged object recognition rates on ETH80.

| Evaluated Methods | 2 train Acc. (%) | 4 train Acc. (%) | 6 train Acc. (%) | 8 train Acc. (%) |
|---|---|---|---|---|
| LPP | 82.71 | 88.83 | 90.64 | 92.63 |
| IsoP | 85.41 | 89.77 | 90.44 | 91.94 |
| IRPCA | 86.21 | 89.23 | 90.07 | 92.34 |
| I-LSPFC | 87.75 | 89.57 | 91.16 | 92.41 |
| LatLRR | 86.03 | 88.14 | 89.54 | 91.24 |
| rLatLRR | 86.18 | 88.56 | 90.13 | 91.81 |
| FLLRR | 85.58 | 88.71 | 90.58 | 91.59 |
| rLRR | 88.44 | 89.53 | 92.11 | 92.29 |
| **rBDLR** | **90.39** | **91.15** | **92.41** | **93.55** |

the number of training data increases. rBDLR obtains enhanced results than the other compared methods.

Table 4: Averaged handwritten recognition rates on USPS.

| Evaluated Methods | 10 train Acc. (%) | 20 train Acc. (%) | 30 train Acc. (%) | 40 train Acc. (%) |
|---|---|---|---|---|
| LPP | 70.17 | 77.60 | 84.49 | 86.10 |
| IsoP | 71.25 | 76.11 | 83.93 | 86.51 |
| IRPCA | 76.88 | 80.98 | 82.15 | 84.89 |
| I-LSPFC | 78.53 | 83.77 | 85.39 | 87.29 |
| LatLRR | 79.19 | 82.23 | 84.34 | 87.58 |
| rLatLRR | 80.94 | 82.87 | 87.33 | 89.03 |
| FLLRR | 79.94 | 83.71 | 88.59 | 90.11 |
| rLRR | 80.27 | 84.72 | 87.66 | 88.09 |
| **rBDLR** | **82.45** | **86.64** | **90.20** | **91.32** |

**Recognition results on COIL100 database.** We evaluate each method for object recognition under various numbers of training samples using COIL100. This database has 7200 images from 100 objects placed on a motorized turntable against a black background. We use the principal components features extracted by *Principal Component Analysis* (PCA) [30] and the dimension is 400. We train on 5, 10, 15 and 20 samples per object class and the recognition results are shown in Table 2. We find that rBDLR performs the best. Specifically, the superiority over the others is more obvious when the training size is relatively small.

**Recognition results on ETH80 database.** We evaluate each method for object recognition on the ETH80 object database. For object recognition, we randomly select 2, 4, 6 and 8 images from each category as training set and test on the rest. The averaged results of each algorithm over 10 runs are shown in Table 3. We find that: (1) the increasing number of training data improves the result of each method; (2) our rBDLR performs the best among all the compared methods, followed by rLRR and I-LSPFC.

**Recognition results on USPS database.** This digit database has 9298 handwritten digits ('0'-'9') and each sample is flattened to a 256 dimensional vector. We randomly select 10, 20, 30 and 40 images per class for training and test on the rest. We show the averaged results in Table 4. We find that rBDLR outperforms the other methods. FLLRR and rBDLR are comparable with each other and are superior to the other methods in most cases.

## 5.5 Image Recognition against Corruptions

We explore the robustness property of each method. UMIST and USPS databases are evaluated. To corrupt data, random Gaussian noise with the fixed variance is added. The number of training images per class is set to 10 for UMIST and USPS. The noisy recognition results are shown in Fig.7, where variance is set to 100, 200, ... , 500, and some examples of noisy images are also illustrated. We find that: (1) the results are decreased when the noise level is increased; (2) our rBDLR delivers higher accuracies than other methods under different noise levels. Specifically, rBDLR degrades slower than other methods with the increasing of variance, which means that our rBDLR method is less sensitive to the noise and corruptions than other methods.

## 5.6 Quantitative Clustering Evaluations

We evaluate each method for clustering images. Two databases, i.e., UMIST face and USPS digits, are used. The clustering task is conducted by performing the K-means clustering algorithm with cosine distance on the recovered or reconstructed data of all the samples for each model. Specifically, for each fixed number K of clusters, we choose K categories from the dataset randomly and use K-category data to form the data matrix for representation learning. For each setting, the results are averaged based on 30 random initialization for K-means. Clustering accuracy (AC) and F-measure are used as evaluation metrics. The values of AC and F-measure on the two databases are described in Table 5 and 6. We find that: (1) generally speaking, the clustering accuracy of each model goes down as the number of categories increases, because clustering more data is usually difficult than clustering less. (2) rBDLR delivers better results than competitors, implying that rBDLR obtains effective data representations for clustering.

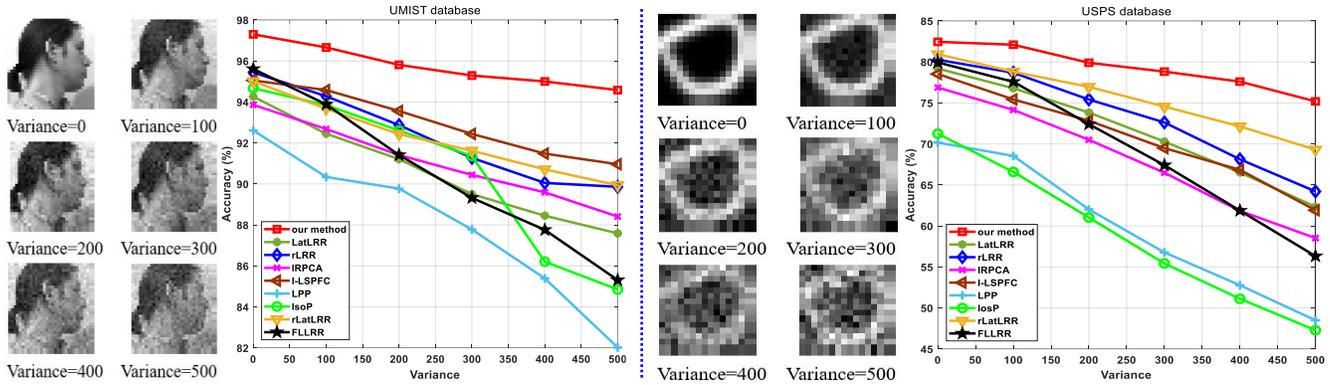

Figure 7: Classification performance vs. varying variance on UMIST (left) and USPS (right) databases.

Table 5: Numerical clustering evaluation results on the UMIST face database.

| Method | Clustering Accuracy ±standard deviation (%) | | | | F-Score values ±standard deviation (%) | | | |
|---|---|---|---|---|---|---|---|---|
| | K=2 | K=4 | K=6 | K=8 | K=2 | K=4 | K=6 | K=8 |
| K-means | 80.2±3.5 | 55.3±5.1 | 50.1±3.1 | 51.1±3.9 | 62.8±2.9 | 45.0±4.3 | 42.7±3.3 | 41.3±4.1 |
| RPCA | 88.5±6.8 | 71.5±10.2 | 62.0±5.6 | 56.3±5.5 | 81.7±19.3 | 62.1±9.4 | 51.7±6.1 | 42.2±4.5 |
| LRR | 88.5±6.0 | 73.5±10.5 | 61.5±8.4 | 52.0±5.8 | 83.3±17.9 | 66.3±8.6 | 51.2±4.4 | 41.7±5.6 |
| IRPCA | 88.6±5.9 | 72.0±6.7 | 60.8±7.0 | 55.1±6.7 | 82.7±18.5 | 64.1±4.4 | 52.8±6.8 | 42.1±4.6 |
| I-LSPFC | 91.0±5.2 | 74.3±11.0 | 61.2±7.0 | 55.9±4.9 | 87.9±13.0 | 65.1±11.4 | 51.7±4.5 | 45.7±4.9 |
| LatLRR | 90.1±5.5 | 72.0±15.5 | 61.8±8.1 | 55.8±8.0 | 85.4±18.9 | 65.4±14.4 | 52.7±5.0 | 43.3±7.0 |
| rLatLRR | 90.5±5.6 | 78.3±7.0 | 61.2±6.2 | 56.1±4.5 | 83.5±22.5 | 66.4±9.3 | 53.6±5.0 | 43.9±5.1 |
| FLLRR | 88.4±8.4 | 78.4±14.4 | 61.0±5.2 | 55.5±4.6 | 84.9±14.8 | 65.2±13.6 | 52.7±3.7 | 42.3±4.2 |
| rLRR | 89.9±11.1 | 71.3±12.2 | 61.3±14.1 | 55.4±5.7 | 80.0±17.8 | 60.5±9.7 | 53.4±11.8 | 46.7±7.3 |
| BDR | 93.4±5.3 | 77.9±13.1 | 58.3±4.5 | 56.2±4.0 | 83.6±17.6 | 67.1±11.9 | 52.7±4.4 | 41.8±4.2 |
| **rBDLR** | **94.5±4.3** | **80.5±11.6** | **64.2±6.7** | **58.8±5.0** | **88.6±14.5** | **70.8±9.4** | **55.1±5.3** | **44.9±4.7** |

Table 6: Numerical clustering evaluation results on USPS handwritten digit database.

| Method | Clustering Accuracy ±standard deviation (%) | | | | F-Score values ±standard deviation (%) | | | |
|---|---|---|---|---|---|---|---|---|
| | K=2 | K=4 | K=6 | K=8 | K=2 | K=4 | K=6 | K=8 |
| K-means | 88.3±13.4 | 79.4±11.4 | 63.4±5.5 | 57.6±7.2 | 83.3±7.1 | 78.4±7.5 | 58.4±5.3 | 46.9±5.3 |
| RPCA | 93.2±1.2 | 83.5±16.2 | 63.7±7.8 | 58.4±4.1 | 86.9±2.1 | 80.7±10.9 | 62.4±6.4 | 48.6±3.4 |
| LRR | 93.0±4.8 | 82.3±11.5 | 63.4±4.7 | 58.4±4.1 | 86.5±3.7 | 81.8±7.7 | 59.3±5.5 | 48.3±2.1 |
| IRPCA | 88.5±12.6 | 83.3±10.8 | 65.5±6.7 | 56.3±4.6 | 83.4±6.7 | 80.6±10.2 | 61.2±5.8 | 49.7±6.0 |
| I-LSPFC | 94.7±0.8 | 84.1±13.5 | 65.7±7.2 | 58.2±8.3 | 87.6±1.5 | 80.8±9.9 | 63.4±6.0 | 50.4±3.5 |
| LatLRR | 92.8±12.7 | 84.5±13.3 | 63.1±7.7 | 57.4±5.4 | 84.9±7.0 | 82.8±70.9 | 61.1±5.8 | 48.8±7.1 |
| rLatLRR | 94.8±2.1 | 86.1±14.7 | 66.2±5.8 | 61.3±6.2 | 88.5±4.6 | 82.7±12.8 | 62.6±3.9 | 52.6±5.9 |
| FLLRR | 93.7±12.4 | 85.8±2.7 | 65.9±5.5 | 60.6±7.0 | 86.7±9.2 | 80.7±4.3 | 62.9±6.5 | 49.4±7.2 |
| rLRR | 92.7±9.4 | 82.3±12.5 | 65.2±8.4 | 58.7±5.9 | 86.4±5.4 | 80.5±6.9 | 61.8±7.7 | 46.7±8.2 |
| BDR | 93.2±1.2 | 84.8±14.9 | 65.2±6.5 | 60.3±6.2 | 86.9±2.2 | 82.0±11.2 | 62.3±5.4 | 50.9±7.6 |
| **rBDLR** | **95.2±0.9** | **86.6±9.2** | **68.3±7.5** | **63.0±7.2** | **90.6±1.7** | **84.5±7.7** | **64.2±6.4** | **52.9±7.1** |

## 6 CONCLUSION AND FUTURE WORK

We have proposed a novel unsupervised representation learning model, i.e., Robust Block-Diagonal Adaptive Locality-constrained Latent Representation (rBDLR). Our rBDLR performs the latent representation under an adaptive locality-based block-diagonal matrix based regularizer in a recovered clean data space, which enables our rBDLR to recover multiple subspaces and extract the adaptive locality-preserving salient features jointly in a robust manner. By using the adaptive locality to define block-diagonal regularizer and using a flexible regularization, we can obtain the block-diagonal coefficients and interpretable subspaces.

We have examined the effectiveness of our rBDLR on several public databases for representation and recognition. The results demonstrate the superior performance of our method over other baselines. Visual image analysis shows the obtained coefficients have block-diagonal structures. In future, we will explore how to extend rBDLR to semi-supervised scenario using labeled and unlabeled data. Extending our rBDLR to handle other real tasks, e.g., background modeling and video analysis, will be explored.

## ACKNOWLEDGMENT

The authors express sincere thanks to reviewers for insightful comments, making this manuscript a higher standard. This work is partially supported by National Natural Science Foundation of China (61672365, 61732008, 61725203 and 61871444) and the Fundamental Research Funds for Central Universities of China (JZ2019HGPA0102). Dr. Zhao Zhang is corresponding author.